\documentclass[11pt]{article}

\usepackage[final]{acl}
\usepackage{tcolorbox}
\usepackage{times}
\usepackage{latexsym}

\usepackage[T1]{fontenc}

\usepackage[utf8]{inputenc}

\usepackage{microtype}

\usepackage{inconsolata}

\usepackage{graphicx}
\usepackage{bm}
\usepackage{multirow}
\usepackage{pifont}
\usepackage{algorithm}
\usepackage{algorithmic}
\usepackage{wrapfig}
\usepackage{colortbl}

\usepackage{amsmath}
\usepackage{amssymb}
\usepackage{mathtools}
\usepackage{amsthm}
\usepackage{graphicx}
\usepackage{hyperref,xcolor}
\usepackage{booktabs}
\usepackage{fontawesome5}

\newcommand{\methodname}{AgentOCR}

\definecolor{HeaderGray}{RGB}{245,245,245}
\definecolor{BlockGray}{RGB}{250,250,250}
\definecolor{OursBlue}{RGB}{234,244,255}
\definecolor{TokenOrange}{RGB}{255,245,235}
\definecolor{DeepRed}{RGB}{165,42,42}
\definecolor{DeepPurple}{RGB}{63,0,95} 
\definecolor{DeepGreen}{RGB}{0,160,80}
\definecolor{DeepBlue}{RGB}{30,60,180}
\definecolor{brown}{RGB}{139,69,19}
\definecolor{DeepOrange}{RGB}{204,85,0}

\title{AgentOCR}
\title{AgentOCR: Reimagining Agent History via Optical Self-Compression}

\author{
 \textbf{Lang Feng\textsuperscript{1,}}\thanks{Equal Contribution
},
 \textbf{Fuchao Yang\textsuperscript{1,}}\footnotemark[1],
 \textbf{Feng Chen\textsuperscript{1}},
 \textbf{Xin Cheng\textsuperscript{1}},
 \textbf{Haiyang Xu\textsuperscript{2}},\\
 \textbf{Zhenglin Wan\textsuperscript{1}},
 \textbf{Ming Yan\textsuperscript{2}},
 \textbf{Bo An\textsuperscript{1,}}\thanks{Corresponding author}
\\
\textsuperscript{1}Nanyang Technological University, Singapore
\\
\textsuperscript{2}Tongyi Lab, Alibaba Group
\\
\texttt{\{lang005,fuchao001\}@e.ntu.edu.sg, boan@ntu.edu.sg} \\
}

\begin{document}
\maketitle
\begin{abstract}
Recent advances in large language models (LLMs) enable agentic systems trained with reinforcement learning (RL) over multi-turn interaction, but practical deployment is bottlenecked by rapidly growing textual histories that inflate token and memory costs. We introduce AgentOCR, a framework that exploits visual tokens' superior information density by representing the accumulated observation-action history as a compact rendered image. To make multi-turn rollouts scalable, AgentOCR proposes segment optical caching. By decomposing history into hashable segments and maintaining a visual cache, this mechanism eliminates redundant re-rendering. Beyond fixed rendering, AgentOCR introduces agentic self-compression, where the agent actively emits a compression rate and is trained with compression-aware reward to adaptively balance task success and token efficiency. We conduct extensive experiments on challenging agentic benchmarks, ALFWorld and search-based QA. Remarkably, AgentOCR preserves over 95\% of text-based agent performance while substantially reducing token consumption (>50\%), yielding consistent token and memory efficiency. Further analysis validates a 20$\times$ rendering speedup from optical caching and effective self-compression balancing. Our code is available at \url{https://github.com/langfengQ/AgentOCR}.
\end{abstract}

\section{Introduction}

\begin{figure*}[t]
    \centering
    \includegraphics[width=0.96\linewidth]{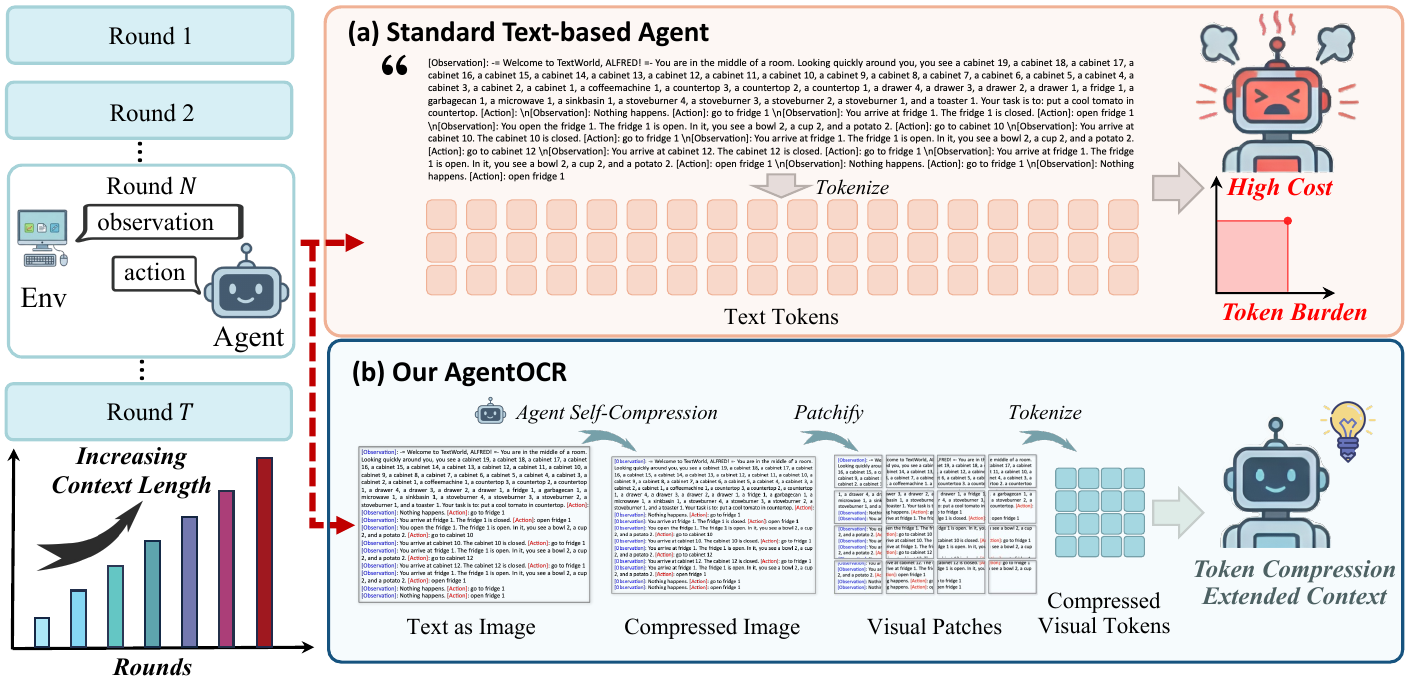}
    \caption{Comparison of text agent and \methodname{}. (\textbf{a}) Text agent accumulates a heavy token burden from raw text history. (\textbf{b}) Our \methodname{} requires significantly fewer visual tokens via optical self-compression.}
    \label{fig:introduciton}
\end{figure*}

Recent advancements in large language models (LLMs)~\citep{gpt,Deepseek-r1,Qwen3,Gemini} have enabled agentic systems~\citep{androidlab,changetal2025main,zhangetal2025belle,xieetal2025gui} that tightly couple perception, deliberation, and actuation, thereby reducing the need for continuous human supervision~\citep{yao2023react,landscape}. These successes increasingly motivate framing agent improvement as reinforcement learning (RL) over long-horizon trajectories-optimizing tool use, planning, and control policies end-to-end from interaction feedback~\citep{feng2025group,wang2025ragen}.

However, RL training for LLM agents remain difficult. A major challenge arises from the burden of \emph{long-context processing}~\citep{long_context1,long_context2}. As agents interact with environments through \emph{multi-turn} decision loops, they must buffer a comprehensive trajectory of past observations and action sequences. As shown in Fig. \ref{fig:introduciton}(a), this historical data accumulates relentlessly, causing the input context to swell to a massive volume of tokens within a single trajectory. Such rapid expansion not only exhausts the finite token budget of current LLMs but also incurs prohibitive inference latency and compute cost due to expensive attention prefill and KV-cache management~\citep{shah2024flashattention,jiang2024minference}.

Recent breakthroughs in vision language models (VLMs)~\citep{bai2025qwen2,Qwen3-VL,InternVL} and optical character recognition (OCR)~\citep{deepseekocr,Paddleocr,xing2025vision} suggest a promising solution: \emph{visual information density}. Notably, DeepSeek-OCR~\citep{deepseekocr} shows that the visual modality can serve as a far more compact carrier of information than text. By rendering textual content into images, the token footprint can be compressed by approximately 10$\times$ compared to raw text tokens, substantially reducing the number of tokens processed during model inference.

Building on this, we propose \textit{\methodname{}}, a visually-grounded method that reimagines agent history not as a string of text, but as a dynamic sequence of images (Fig.~\ref{fig:introduciton}(b)). Specifically, \methodname{} represents the accumulated observation-action history as a compact image and conditions the agent’s policy on this visual history for the multi-turn decision making. To ensure scalability and efficiency in long rollouts, \methodname{} introduces a \textit{segment optical caching}. This mechanism decomposes the history context into segments and maintains a hash-based optical cache, allowing the agent to reuse previously rendered content and avoid redundant processing as the history expands. Beyond static image, \methodname{} features \textit{agentic self-compression}, which empowers the agent to actively modulate its own visual fidelity at each step. By learning via RL, \methodname{} adaptively selects the most appropriate compression factor to save token costs, thereby achieving a favorable balance between task success and efficiency.

We evaluate \methodname{} on two challenging agentic benchmarks: ALFWorld~\citep{ALFWorld}, which features long-horizon decision-making, and search-based QA~\citep{jin2025search}, characterized by highly text-dense interactions. Our results demonstrate that \methodname{} preserves over 95\% of the task performance of strong text-based agent pipelines, while reducing token consumption by over 50\% (up to 80\% in peak tokens), leading to substantially lower overhead. Furthermore, our analysis validates that segment optical caching accelerates rendering by over 20$\times$, and self-compression mechanism effectively optimizes the trade-off between information density and cost.

\section{Related Work}
\subsection{Reinforcement Learning for LLM Agents}
RL~\citep{sutton2018reinforcement} has become a commonly adopted paradigm for aligning agents with human preferences~\citep{Stiennon2020,ouyang2022training,Rafailov2023} and improving their behaviors~\citep{sheng2024hybridflow,wang2025reinforcement} in complex scenarios, such as PPO~\citep{schulman2017proximalpolicyoptimizationalgorithms}, GRPO~\citep{shao2024deepseekmath}, Dr. GRPO~\citep{liu2025understandingr1zeroliketrainingcritical}, Clip-Cov~\citep{cui2025entropymechanismreinforcementlearning}, GSPO~\citep{zheng2025groupsequencepolicyoptimization}, DAPO~\citep{yu2025dapoopensourcellmreinforcement}, GiGPO~\citep{feng2025group}, and HGPO~\citep{he2026hierarchy}. These RL-trained agents have been widely deployed in dynamic and open-ended environments, spanning interactive games~\citep{narasimhan,brockman2016openaigym}, GUI control~\citep{10.5555,ye2025mobileagentv3fundamentalagentsgui,liu2025pcagenthierarchicalmultiagentcollaboration}, embodied tasks~\citep{ALFWorld}, as well as web~\citep{ArCHer,putta2024agentqadvancedreasoning,feng2025towards} and tool-enhanced environments~\citep{qian2025toolrl,sun2025zerosearchincentivizesearchcapability,dong2025agentic,xue2025simpletirendtoendreinforcementlearning}.
\subsection{Optical Character Recognition}
OCR converts textual information in images into computer-readable text and has been widely applied in document digitization~\citep{Doermann,4376991,article}, image text extraction~\citep{s11263,baek2019wrongscenetextrecognition,long2020texts}, and document parsing~\citep{katti-etal-2018-chargrid,LayoutLM,DocParser}. With the advent of deep learning, OCR systems have shifted toward end-to-end frameworks, including Nougat~\citep{nougat}, Donut~\citep{kim2022donut}, TrOCR~\citep{TrOCR}, Pix2Struct~\citep{Pix2Struct} and GOT-OCR2.0~\citep{GOT-OCR}, enabling unified image-to-text modeling for complex scenarios. Recently, OCR has begun to be explored as a vision-text compression mechanism, such as DeepSeek-OCR~\citep{deepseekocr}, VIST~\citep{xing2025vision}, and Glyph~\citep{cheng2025glyphscalingcontextwindows}. These approaches offer a novel solution for processing extremely long contexts, however, research in this direction remains in its early stages.

\subsection{Agent Memory}
Long-horizon interaction with LLM-based agents requires persistent memory, but naively appending full histories quickly exceeds fixed context windows~\citep{memory1,landscape,hu2025memory}. Accordingly, long-context language modeling focuses on efficiency and length generalization, including sparse or hierarchical attention~\citep{beltagy2020longformer,fu2024moa,xiao2024duoattention}, recency-biased positional schemes~\citep{ding2024longrope,su2024roformer,xiong2024effective}, and prompt compression~\citep{ge2023context,yoon2024compact,zhang2024adacomp}. In parallel, retrieval-based methods treat external stores as non-parametric memory and fetch relevant information on demand~\citep{ge2023context,chen2024bge}. Beyond context optimization, recent agent frameworks introduce explicit memory modules to support long-term behavior, ranging from virtualized context management and scalable backends~\citep{packer2023memgpt,chhikara2025mem0} to structured or hierarchical representations for extended and multi-agent tasks~\citep{xu2025mem,anokhin2024arigraph,hu2025hiagent,zhang2025g}. More recent work explores learning-based memory control, where agents adaptively write, retain, and retrieve information, treating memory as an active component~\citep{yan2025memoryr1,zhou2025mem1,yu2025memagent}.

\section{Preliminaries}
\label{sec:background}

\subsection{Problem Setup}
We formulate the interaction between an LLM agent and an environment (e.g., a physical simulator or an external tool API) as a sequential decision-making process over a finite horizon $T \in \mathbb{N}$. 
The agent is instantiated as an LLM parameterized by $\theta$, and its behavior is modeled as a stochastic policy $\pi_\theta$. At each step $t \in \{1,\dots,T\}$, the agent receives an observation $\mathbf{o}_t \in \mathcal{O}$ (e.g., API outputs) and has the interaction history up to step $t$ as 
\begin{equation}
\mathbf{h}_t = (\mathbf{o}_1, \mathbf{a}_1, \mathbf{o}_2, \mathbf{a}_2, \dots, \mathbf{o}_t).
\end{equation}
The agent then samples a textual action $\mathbf{a}_t \sim \pi_\theta(\cdot \mid \mathcal{I}, \mathbf{h}_t)$, where $\mathcal{I}$ denotes the task instruction, $\mathbf{a}_t \in \mathcal{V}^n$ is a token sequence drawn from the vocabulary $\mathcal{V}$ with maximum length $n \in \mathbb{N}$. The action $\mathbf{a}_t$ is flexible and may explicitly include intermediate reasoning (e.g., chain-of-thought~\citep{wei2022chain}) or tool invocations. After executing $\mathbf{a}_t$, the environment returns a scalar reward $r_t$ and the next observation $\mathbf{o}_{t+1}$. Notably, $\mathbf{h}_t$ often becomes extensively long due to extended horizons or verbose environmental observations, posing significant challenges for context processing.

\subsection{Agentic Reinforcement Learning}
RL has become a standard post-training paradigm for enhancing LLM agents. 
While our work is algorithm-agnostic and compatible with various agentic RL algorithms, we consider Group Relative Policy Optimization (GRPO)~\citep{shao2024deepseekmath} as the representative algorithm due to its simplicity and efficiency.

Specifically, GRPO samples a group of trajectories $\{\tau_i\}_{i=1}^G$ for each input and estimates advantages $\hat{A}_i$ by normalizing rewards within the group. To ensure training stability, we optimize the clipped surrogate objective:
\begin{equation}
\small
J(\theta)=\mathbb{E} \Big[ \frac{1}{GT} \sum_{i=1}^G \sum_{t=1}^{T} \min \Big( \rho_{t,i} \hat{A}_i, \text{clip}(\rho_{t,i}, 1 \pm \epsilon) \hat{A}_i \Big) \Big],
\label{eq:grpo_objective}
\end{equation}
where $\rho_{t,i} = \frac{\pi_\theta(\mathbf{a}_{t,i} \mid \mathcal{I}, \mathbf{h}_{t,i})}{\pi_{\theta_{\text{old}}}(\mathbf{a}_{t,i} \mid \mathcal{I}, \mathbf{h}_{t,i})}$ is the importance sampling ratio, $\epsilon$ is the clipping hyperparameter. Here, we omit the KL-divergence regularization for notational brevity. Crucially, optimizing Eq.~(\ref{eq:grpo_objective}) necessitates computing gradients over the entire cumulative history $\mathbf{h}_t$. In realistic agent scenarios, $\mathbf{h}_t$ rapidly accumulates thousands of tokens (e.g., >10k tokens in multi-turn search tasks~\citep{jin2025search}). Since the computational complexity and memory footprint scale with token count, processing such lengthy textual trajectories becomes prohibitively expensive.

\section{\methodname{}}

Agentic tasks require the policy to condition on an \textit{ever-growing multi-turn interaction history}. This growth creates a severe bottleneck in practice. Not only does it reach the model’s context window limits, but it also drives up computational costs as transformer computation scales superlinearly with sequence length.
Hence, efficient token compression is imperative for deploying LLM agents~\citep{kong2025token}.

\begin{figure*}
\centering
\includegraphics[width=0.96\linewidth]{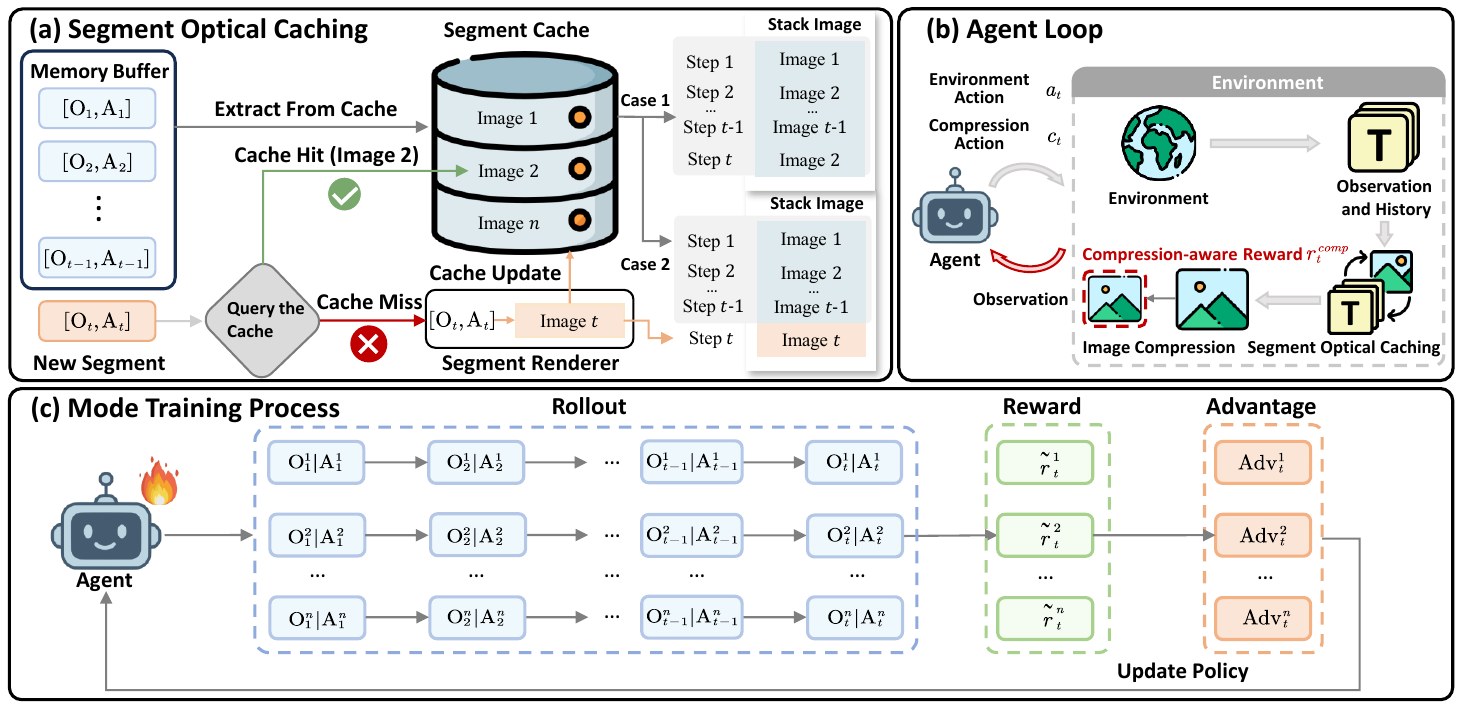}
\caption{Overview of \methodname{}.
\textbf{(a)} Segment optical caching decomposes the history context into segments, reuses cached renderings via content keys, and assembles the optical memory by stacking segment images.
\textbf{(b)} The agent receives the optical observation and history, selects an environment action, and a compression rate.
\textbf{(c)} The agent is trained with RL, jointly optimizing task performance and token efficiency.
}
\label{fig:overview}
\end{figure*}

In this section, we introduce \methodname{} that addresses this bottleneck by reimagining interaction history as an \textit{optical memory}. Instead of processing raw textual logs, we render the accumulated history into a compact visual representation. By leveraging the superior information density of visual tokens compared to text, this approach substantially reduces the token footprint while maintaining full access to historical details. To ensure scalability and dynamic adaptability in long-horizon rollouts, \methodname{} incorporates two innovations: (\textbf{1}) \textit{segment optical caching} (Fig.~\ref{fig:overview}(a)), which eliminates redundant rendering overhead by systematically reusing cached visual segments, and (\textbf{2}) \textit{agentic self-compression} (Fig.~\ref{fig:overview}(b)), a mechanism that empowers the agent to actively modulate the compression rate, thereby optimizing the trade-off between information density and token cost.

In the remainder of this section, we detail optical encoding in Sec.~\ref{sec:agentocr:optical} for visual representation, followed by segment optical caching in Sec.~\ref{sec:agentocr:cache} and self-compression and RL training in Sec.~\ref{sec:agentocr:self-compression}.

\subsection{Optical Memory Encoding}
\label{sec:agentocr:optical}
\paragraph{Memory buffer.}
\methodname{} maintains an external memory buffer $\mathcal{M}_t$ that stores the interaction records up to step $t$.
Each record contains an observation-action pair $(\mathbf{o}_t,\mathbf{a}_t)$ (or task-specific equivalents such as tool queries and results).
The memory module serializes the entire interaction history into a textual form $\mathbf{h}_t = \mathrm{Fetch}(\mathcal{M}_{t-1})$.

\paragraph{Memory rendering.}
We define a deterministic renderer $\mathcal{R}$ that maps the textual interaction history to an RGB image
$\mathbf{I}_t = \mathcal{R}(\mathbf{h}_t;\psi)$,
where $\psi$ denotes rendering hyperparameters (e.g., font family and size, colors, padding, and bounds on image width and height).
At step $t$, \methodname{} constructs a multimodal policy input by combining $\mathcal{I}$ and the rendered history image $\mathbf{I}_t$, and samples an action from a vision-language policy: $\mathbf{a}_t \sim \pi_\theta(\cdot \mid \mathcal{I}, \mathbf{I}_t)$.
The sampled action $\mathbf{a}_t$ is then applied to the environment to obtain the next observation $\mathbf{o}_{t+1}$, and the memory buffer is updated accordingly.

\subsection{Segment Optical Caching}
\label{sec:agentocr:cache}
Rendering the entire history $\mathbf{h}_t$ from scratch at every step is wasteful and becomes a major latency bottleneck in multi-turn rollouts.
A naive alternative is to render only the newly appended context and append it to the previously rendered history image.
This yields near-constant per-step rendering overhead, but it cannot reuse recurring content, and its memory still grows with the accumulated rendered pixels.

\methodname{} instead performs caching at the granularity of segments.
The core idea is to decompose the full history context into independent segments and memory rendered segments in a dictionary keyed by segment content.
At each step, we assemble the history image by stacking cached segment images in order, rendering only segments that have not been seen before.
This cache naturally accelerates both recurring boilerplate and repeated tool outputs, and can also reuse newly arrived observations/actions whenever they match previously observed segments.

\paragraph{Segment representation.}
We split the history context into segments.
Let $\mathrm{Split}(\mathbf{h})=(\ell_1,\ldots,\ell_K)$ denote this operation, where each $\ell_i$ is a text segment.
We use a deterministic segment renderer $\mathcal{R}(\ell;\psi)$ that maps a single segment to an RGB image under the same rendering hyperparameters.

\paragraph{Segment cache.}
For each environment instance $e$, \methodname{} maintains a per-episode cache
\begin{equation}
\mathcal{C}^{(e)}=\{(k(\ell), \mathbf{I}(\ell))\},
\end{equation}
where $k(\ell)$ is a fast content key (e.g., a hash of the normalized segment text, optionally including style metadata) and $\mathbf{I}(\ell)$ is the rendered image of segment $\ell$.
Unlike naive cache, which incrementally renders the newly appended context, this cache stores each unique segment at most once and reuses it whenever the same segment reappears.
In our implementation, $\mathcal{C}^{(e)}$ persists within an episode and is reset at episode boundaries.

\paragraph{Cache lookup and assembly.}
At step $t$, we obtain the full history $\mathbf{h}_t$ and split it into segments
$\mathrm{Split}(\mathbf{h}_t)=(\ell_{t,1},\ldots,\ell_{t,K_t})$.
For each segment $\ell_{t,i}$, we first query the cache. On a miss, we render and insert it:
\begin{equation}
\mathbf{I}(\ell_{t,i})=
\begin{cases}
\mathcal{C}^{(e)}[k(\ell_{t,i})], & \text{if}\; k(\ell_{t,i})\in\mathcal{C}^{(e)},\\
\mathcal{R}(\ell_{t,i};\psi), & \text{otherwise}.
\end{cases}
\end{equation}
\begin{equation}
\text{if miss:}\quad
\mathcal{C}^{(e)}[k(\ell_{t,i})]\leftarrow \mathbf{I}(\ell_{t,i}).
\end{equation}
The full optical memory image is then constructed by vertically stacking segment images in order:
\begin{equation}
\mathbf{I}_t=\mathrm{Stack}\Big(\mathbf{I}(\ell_{t,i})\Big)_{i=1}^{K_t}.
\end{equation}
Because newly appended observations or actions are processed as just additional segments, they can be reused without re-rendering whenever they match cached content (e.g., repeated queries and tool responses).

\paragraph{Complexity.}
Let $U_t$ be the number of cache-miss segments among $\{\ell_{t,i}\}_{i=1}^{K_t}$.
The per-step rendering cost becomes $O(U_t)$ segment renders, while cache hits require only dictionary lookup and image stacking.
In many agent workloads, interaction logs contain substantial repetition, so typically $U_t \ll K_t$ and rendering overhead is significantly reduced.
For spatial complexity, the cache stores one image per unique segment per episode, yielding $O(|\mathcal{C}^{(e)}|)$ images rather than $O(T)$ full-history images, thereby avoiding heavy duplication across environment timesteps.

\subsection{Agentic Self-Compression}
\label{sec:agentocr:self-compression}
\paragraph{Compression decision.}
Instead of treating the optical renderer $\mathcal{R}$ (defined in Sec.~\ref{sec:agentocr:optical}) as a static background process, \methodname{} exposes it as an executable tool.
We conceptualize this interaction as a parameterized invocation alongside environment actions, the policy generates a structured call via <compression>$c_t$</compression>, where the compression factor $c_t \ge 1$ dynamically modulates the rendering fidelity.
This design aligns with standard tool-use paradigms, allowing the agent to explicitly query its interaction history with variable precision.
Upon receiving the call, the system executes the renderer with the specified compression factor.
Formally, this produces a scaled image $\mathbf{I}_{t+1}$:
\begin{equation}
\label{eq:compress}
\mathrm{size}(\mathbf{I}_{t+1}) = 
\left(
\left\lfloor \tfrac{H_{t+1}}{\sqrt{c_t}} \right\rfloor,
\left\lfloor \tfrac{W_{t+1}}{\sqrt{c_t}} \right\rfloor
\right).
\end{equation}
This spatial downsampling operation effectively reduces the number of visual tokens. Consequently, the agent can strategically modulate the compression rate based on specific task characteristics, thereby optimizing the trade-off between token cost and information density.

\paragraph{Compression-aware reward.}
To incentivize the agent to identify suitable compression without compromising task success, we introduce a compression-aware reward term for RL training. This reward is strictly conditioned on episode success, ensuring that the agent treats compression as a secondary cost-optimization objective rather than a primary goal. Then, we employ a logarithmic reward formulation that reflects the diminishing returns of information density. Formally,
let $\mathbb{I}_{\text{succ}}(\tau)\in\{0,1\}$ be the success indicator.
The compression reward at step $t$ is defined as:
\begin{equation}
\label{eq:compress_reward}
    r^{\text{comp}}_t =
    \begin{cases}
    \ln(c_t), & \text{if} \;  \mathbb{I}_{\text{succ}}(\tau)=1,\\
    0, & \text{otherwise}.
    \end{cases}
\end{equation}
The total reward used for RL optimization is given by $\widetilde{r}_t = r_t + \lambda r^{\text{comp}}_t$,
where $\lambda \ge 0$ is the weight parameter of compression reward, which controls the trade-off between task performance and compression efficiency.
This scalar reward is used by the agentic RL optimizer (e.g., GRPO) to update the policy $\pi_\theta$ through the objective in Eq.~(\ref{eq:grpo_objective}). 
However, applying the compression reward $r^{\text{comp}}_t$ at every training iteration can induce overly greedy behavior, where the agent aggressively increases compression to maximize immediate reward.
To mitigate this effect, we adopt an intermittent reinforcement schedule, injecting the compression reward only at intervals of $K$ training iterations. This schedule introduces the efficiency signal periodically while maintaining the primary optimization pressure on task completion. Under this design, \methodname{} learns to allocate vision tokens adaptively, achieving further token reduction while maintaining strong task performance.

\section{Experiment}
In this section, we present comprehensive empirical evaluations of \methodname{} across two representative multi-turn agent benchmarks. Specifically, our experiments aim to investigate the following key aspects:
(\textbf{1}) the comparative performance and token efficiency of \methodname{} relative to text-based agents; 
(\textbf{2}) the quantitative analysis of the vision-text compression ratio of \methodname{}; 
(\textbf{3}) the computational efficiency of the segment optical caching mechanism; 
(\textbf{4}) the ablation study on the effectiveness of the self-compression mechanism.

\subsection{Experimental Setup}
\paragraph{Benchmarks.}
We evaluate \methodname{} on ALFWorld~\citep{ALFWorld} and search-based QA~\citep{jin2025search}. Both benchmarks exhibit sustained context growth but with different interaction profiles. ALFWorld contains the embodied tasks, requiring the agent to manipulate objects within a simulated household environment. Search-based QA, conversely, focuses on multi-turn tool use and information retrieval. The agent must actively interact with a search engine to query external knowledge, requiring it to deal with denser, web-style textual traces.

\paragraph{Baselines.}
We compare \methodname{} against both text-based and optical-history variants across prompting and RL regimes. Specifically, we evaluate: (\textbf{1}) \textit{Text (w/o RL)}, which feeds raw textual history to a text-only model; (\textbf{2}) \textit{OCR (w/o RL)}, which conditions the model on rendered optical memory images without RL; and (\textbf{3}) \textit{Text + GRPO}, a strong baseline that applies RL directly to the raw textual context. In contrast, our \methodname{} applies GRPO to the optical-memory agent, enabling efficient optimization over compact visual histories.

\paragraph{Training details.}
We use the Qwen2.5-VL~\citep{bai2025qwen2} family as the backbone models. Text-only variants use Qwen2.5-3B/7B-Instruct, while optical-history variants use Qwen2.5-VL-3B/7B-Instruct. We keep all training settings and hyperparameters identical across methods to ensure controlled comparisons. For extra hyperparameters of \methodname{}, we set $\lambda=0.01$ and $K=5$. More details are provided in Appendix~\ref{app:experiment}.

\begin{table*}[t]
\centering
\resizebox{1\textwidth}{!}{%
\begin{tabular}{lccccccc cc}
\toprule
\rowcolor{HeaderGray}
\cellcolor{HeaderGray} &
\multicolumn{7}{c}{\textbf{ALFWorld}} &
\multicolumn{2}{c}{\cellcolor{TokenOrange}\textbf{Tokens/Step}} \\
\rowcolor{HeaderGray}
\textbf{Method}
& \textbf{Pick \& Place} & \textbf{\phantom{0}Look\phantom{0}} & \textbf{\phantom{0}Clean\phantom{0}} & \textbf{\phantom{0}Heat\phantom{0}} & \textbf{\phantom{0}Cool\phantom{0}}
& \textbf{Pick2 \& Place} & \textbf{Avg.}
& \cellcolor{TokenOrange}\textbf{Avg.} & \cellcolor{TokenOrange}\textbf{Max.} \\
\midrule

\rowcolor{BlockGray}
\multicolumn{10}{c}{\textit{Qwen2.5-(VL)-3B-Instruct}} \\
    Text (w/o RL)   & 34.7 & 18.4 & 12.7 & 7.3 & 14.5 & 10.4 & 16.3 & \cellcolor{TokenOrange}1.09k & \cellcolor{TokenOrange}3.04k \\
OCR (w/o RL)   & 42.8 & 21.8 & 10.1 & 6.2 & 6.2 & 9.9 & 16.2 & \cellcolor{TokenOrange}0.49k & \cellcolor{TokenOrange}1.63k \\
Text + GRPO    & 92.6 & 85.7 & 70.6 & 86.6 & 79.3 & 65.0 & 79.9 & \cellcolor{TokenOrange}1.02k & \cellcolor{TokenOrange}3.13k \\
\rowcolor{OursBlue}
\textbf{\methodname{}}   & 91.9 & 81.8 & 76.0 & 73.3 & 76.1 & 70.0 & 78.2 & \cellcolor{TokenOrange}0.38k\textsubscript{\textcolor{DeepGreen}{(61.7\%$\downarrow$)}} & \cellcolor{TokenOrange}1.14k\textsubscript{\textcolor{DeepGreen}{(63.6\%$\downarrow$)}} \\

\midrule

\rowcolor{BlockGray}
\multicolumn{10}{c}{\textit{Qwen2.5-(VL)-7B-Instruct}} \\
Text (w/o RL)   & 67.6 & 35.4 & 19.3 & 31.3 & 30.1 & 4.4 & 31.3 & \cellcolor{TokenOrange}1.08k & \cellcolor{TokenOrange}3.36k \\
OCR (w/o RL)    & 61.0 & 33.2 & 17.2 & 11.6 & 12.5 & 16.5 & 25.3 & \cellcolor{TokenOrange}0.47k & \cellcolor{TokenOrange}1.36k \\
Text + GRPO            & 92.6 & 93.8 & 85.2 & 80.0 & 82.7 & 56.5 & 81.8 & \cellcolor{TokenOrange}0.95k & \cellcolor{TokenOrange}2.81k \\
\rowcolor{OursBlue}
\textbf{\methodname{}}      & 95.6 & 96.2 & 78.1 & 73.2 & 72.4 & 72.0 & 81.2 & \cellcolor{TokenOrange}0.43k\textsubscript{\textcolor{DeepGreen}{(54.7\%$\downarrow$)}} & \cellcolor{TokenOrange}1.22k\textsubscript{\textcolor{DeepGreen}{(56.6\%$\downarrow$)}} \\
\bottomrule
\end{tabular}
}
\caption{
Performance on ALFWorld tasks. We report the success rate (\%) and the average and peak memory context token cost per step.
}
\label{tab:main_results_alfworld}
\end{table*}

\begin{table*}[t]
\centering
\setlength{\tabcolsep}{4.5pt}
\renewcommand{\arraystretch}{1.15}
\resizebox{\textwidth}{!}{%
\begin{tabular}{lcccccccccc}
\toprule
\rowcolor{HeaderGray}
\cellcolor{HeaderGray} &
\multicolumn{3}{c}{\textbf{Single-Hop}} &
\multicolumn{5}{c}{\textbf{Multi-Hop}}  &
\multicolumn{2}{c}{\cellcolor{TokenOrange}\textbf{Tokens/Step}} \\
\rowcolor{HeaderGray}
\textbf{Method} &
\textbf{NQ$^\dagger$} & \textbf{TriviaQA$^\star$} & \textbf{PopQA$^\star$} &
\textbf{HotpotQA$^\dagger$} & \textbf{2Wiki$^\star$} & \textbf{MuSiQue$^\star$} & \textbf{Bamboogle$^\star$} & \textbf{Avg.} &
\cellcolor{TokenOrange}\textbf{Avg.} & \cellcolor{TokenOrange}\textbf{Max.} \\
\midrule

\rowcolor{BlockGray}
\multicolumn{11}{c}{\textit{Qwen2.5-(VL)-3B-Instruct}}\\
Text (w/o RL)     & 9.4 & 31.3 & 19.8 & 15.0 & 14.8 & 4.7 & 16.8 & 15.9 & \cellcolor{TokenOrange}0.48k & \cellcolor{TokenOrange}7.34k \\
OCR (w/o RL)      & 10.2 & 27.7 & 10.9 & 9.1 & 12.2 & 3.7 & 15.2 & 12.7  & \cellcolor{TokenOrange}0.15k & \cellcolor{TokenOrange}1.33k \\
Text + GRPO              & 39.3 & 60.6 & 41.1 & 37.4 & 34.6 & 15.4 & 26.4 & 36.4  & \cellcolor{TokenOrange}0.61k & \cellcolor{TokenOrange}9.55k \\
\rowcolor{OursBlue}
\textbf{\methodname{}}        & 38.6 & 56.5 & 41.7 & 33.6 & 30.7 & 14.6 & 24.0 & 34.2  & \cellcolor{TokenOrange}0.26k\textsubscript{\textcolor{DeepGreen}{(57.4\%$\downarrow$)}} & \cellcolor{TokenOrange}2.50k\textsubscript{\textcolor{DeepGreen}{(73.8\%$\downarrow$)}} \\
\midrule

\rowcolor{BlockGray}
\multicolumn{11}{c}{\textit{Qwen2.5-(VL)-7B-Instruct}}\\
Text (w/o RL)     & 10.4 & 32.4 & 22.3 & 15.8 & 15.4 & 7.2 & 19.2 & 17.5  & \cellcolor{TokenOrange}0.70k & \cellcolor{TokenOrange}10.96k \\
OCR (w/o RL)      & 6.9 & 30.4 & 12.0 & 10.5 & 9.1 & 5.5 & 24.0 & 14.0  & \cellcolor{TokenOrange}0.26k & \cellcolor{TokenOrange}2.21k \\
Text + GRPO   & 45.1 & 63.7 & 44.0 & 43.6 & 43.2 & 16.8 & 37.6 & 41.9  & \cellcolor{TokenOrange}0.73k & \cellcolor{TokenOrange}13.84k \\
\rowcolor{OursBlue}
\textbf{\methodname{}}        & 43.1 & 61.0 & 45.4 & 40.8 & 38.3 & 15.7 & 36.8 & 40.1  & \cellcolor{TokenOrange}0.36k\textsubscript{\textcolor{DeepGreen}{(50.7\%$\downarrow$)}} & \cellcolor{TokenOrange}2.65k\textsubscript{\textcolor{DeepGreen}{(80.9\%$\downarrow$)}} \\
\bottomrule
\end{tabular}%
}
\caption{
Performance on search-based QA tasks. We report the exact matching score (\%) and the average and peak memory context token cost per step. $^\dagger$ and $^\star$ denote in-domain and out-of-domain respectively.
}
\label{tab:main_results_qa}
\end{table*}
\subsection{Main Results}
We first evaluate the overall performance of all methods, with results reported in Tab.~\ref{tab:main_results_alfworld} and Tab.~\ref{tab:main_results_qa}.
A direct comparison of the inference-only baselines (``Text'' vs. ``OCR'') reveals the inherent efficiency advantage of the visual modality. Across both benchmarks, optical history drastically reduces token consumption, cutting average usage by approximately 55\% on ALFWorld and 70\% on search tasks. However, this compression initially comes at a cost. The off-the-shelf VLM struggles to ground the condensed visual history effectively, resulting in a substantial performance drop relative to their text-based counterparts.

\methodname{} effectively bridges this gap through RL training, aligning the policy to the visual modality to attain task performance comparable to text-based baselines across model scales. On ALFWorld, \methodname{} with 3B and 7B models achieves 78.2\% and 81.2\% respectively, virtually matching the text agents (within a $\sim$1\% margin). This trend holds for search tasks, where \methodname{} retains over 95\% of the performance of the Text+GRPO baselines (e.g., achieving 40.1\% vs. 41.9\% on the 7B model). Crucially, these results underscore that \methodname{} offers a highly favorable trade-off between token cost and task success. Rather than merely matching the baseline, our method fundamentally alleviates the inference bottleneck by slashing token consumption by >50\% (up to 80.9\% in peak contexts), proving that high-density visual representations can support rigorous agentic reasoning with significantly reduced overhead.

\subsection{Vision-Text Compression Analysis}
\label{sec:compression_analysis}

\begin{figure}[t]
    \centering
    \includegraphics[width=1.0\linewidth]{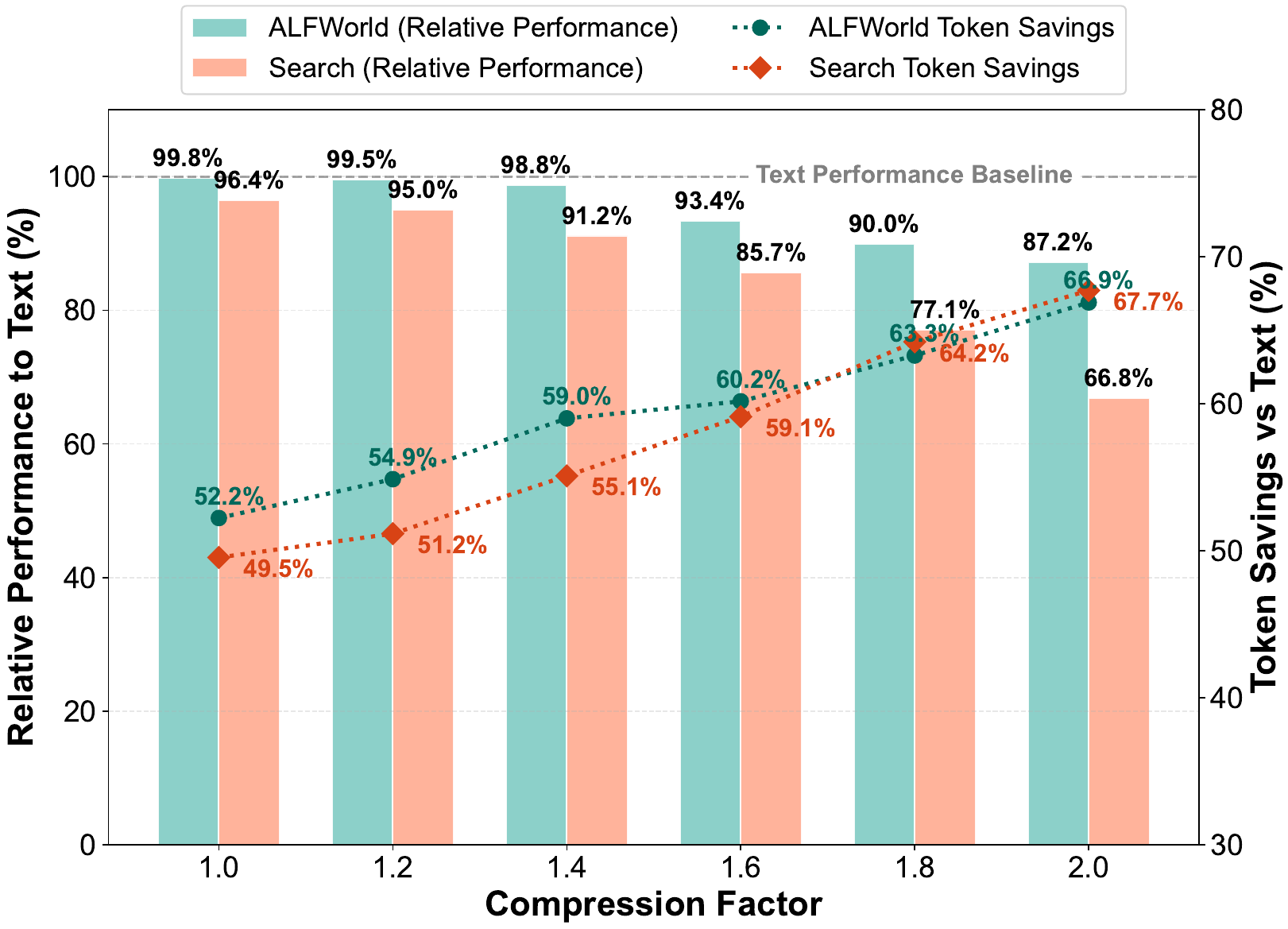}
    \caption{Vision-text compression efficiency. The bars (left axis) denote the success rate relative to the text-based agent baseline, while the lines (right axis) indicate the percentage of tokens saved.}
    \label{fig:compression_tradeoff}
\end{figure}

In this part, we investigate the trade-off between token compression efficiency and task performance using the trained \methodname{} (7B) across varying fixed compression factors ($c_t \in [1.0, 2.0]$). As illustrated in Fig.~\ref{fig:compression_tradeoff}, increasing the compression factor yields substantial gains in token efficiency but incurs a performance penalty. Notably, we identify a robust compression zone up to $\sim$ 55\% token savings (at $c_t=1.2$), where the model successfully maintains over 95\% of the text-based performance (99.5\% for ALFWorld and 95.0\% for Search). However, surpassing this efficiency threshold triggers an accelerated performance decay. For instance, as savings increase further to $\sim$ 67\% ($c_t=2.0$), the average performance drops significantly, highlighting the non-linear tension between information density and reasoning accuracy.

We further observe a distinct divergence in robustness beyond this threshold. ALFWorld demonstrates high resilience, retaining 87.2\% of performance even at $c_t=2.0$, likely due to its reliance on coarse-grained scene understanding. In contrast, the text-dense search task is highly sensitive, with performance plummeting to 66.8\% as aggressive downscaling blurs critical textual cues. These findings underscore the necessity of our \textit{agentic self-compression} mechanism. While a static image offers a safe baseline, a dynamic policy is required to exploit higher compression rates in robust steps while reverting to high fidelity for sensitive reasoning, thereby breaking the static trade-off ceiling.

\subsection{Analysis of Cache}
\label{sec:exp-cache}

To evaluate the scalability of segment optical caching, we compare it against two alternatives: \emph{no cache}, which re-renders the full history $\mathbf{h}_t$ at every step, and \emph{naive cache}, which incrementally renders the newly appended context at step $t$ and appends it to a growing optical-memory image.

\begin{table*}[t]
\centering
\small
\begin{tabular}{l|ccc|ccc}
\toprule
\rowcolor{HeaderGray}
 &
\multicolumn{3}{c|}{\textbf{Render Time}} &
\multicolumn{3}{c}{\textbf{Cache Mem}} \\
\rowcolor{HeaderGray}
\textbf{Method} & \textbf{Avg (ms)} & \textbf{Grow (ms/step)} & \textbf{Speedup}$\uparrow$
& \textbf{Peak (MB)} & \textbf{Grow (MB/step)} & \textbf{Mem Save}$\uparrow$\\
\midrule
No Cache     & 3509.39 & 115.43 & 1.00$\times$   & --      & --     & -- \\
Naive Cache  & 203.08 & 0.03 & 17.28$\times$  & 151.41 & 2.79 & 0.00\% \\
\rowcolor{OursBlue}
\textbf{Ours} & \textbf{168.77} & \textbf{-1.23} & \textbf{20.79$\times$} & \textbf{110.80} & \textbf{1.91} & \textbf{26.82\%} \\
\bottomrule
\end{tabular}
\caption{Cache mechanism ablation. \emph{Growth/step} is the slope from a least-squares linear fit over steps 1--50. \emph{Speedup} is relative to ``no cache''. \emph{Mem Saving} is relative to the peak cache memory of ``naive cache''.}
\label{tab:cache_ablation}
\end{table*}

Tab.~\ref{tab:cache_ablation} shows that \emph{no cache} suffers from large latency and strong growth over time (3509.39 ms on average and 115.43 ms/step), reflecting the redundant cost of repeatedly rendering an increasingly long history. \emph{Naive cache} removes this time growth and achieves a 17.28$\times$ speedup, since each step renders only the appended suffix, yielding an effective per-step cost of $O(1)$ in typical rollouts. However, because it treats each newly arrived line as distinct and permanently appends the rendered result, its cache memory still grows with rollout length (151.41 MB peak and 2.79 MB/step), i.e., space scales with the accumulated rendered pixels.

Segment optical caching further reduces per-step rendering work. By memoizing newline-level segments in a content-keyed dictionary, newly appended context can often be satisfied by cache hits and thus requires no rendering. Therefore, the number of cache-miss segments may shrink as the cache warms up, producing a negative time growth ($-1.23$ ms/step) and the best average latency (168.77 ms). Meanwhile, segment-level reuse also reduces redundant storage (110.80 MB), corresponding to a 26.82\% peak-memory saving relative to \emph{naive cache} and aligning with space scaling dominated by the number of unique segments rather than the number of steps.

\subsection{Analysis of Self-Compression}
\label{sec:exp-self-compression}

\begin{table}[t]
\centering
\resizebox{\linewidth}{!}{
\begin{tabular}{lccc}
\toprule
\rowcolor{HeaderGray}
\textbf{Configuration} 
& \textbf{SR (\%)} 
& \textbf{Avg. $c_t$} 
& \textbf{Avg. Vis. Tok.} \\
\midrule
\rowcolor{BlockGray}\multicolumn{4}{c}{\textit{Without RL}} \\ 
w/o Self-Compression & 12.1 & 1.00 & 441.2 \\
Self-Compression     & 11.8 & 1.05 & 436.9 \\
\midrule
\rowcolor{BlockGray}\multicolumn{4}{c}{\textit{RL Training}} \\ 
w/o Self-Compression       & 78.4 & 1.00 & 458.1 \\
Self-Compression ($K{=1}$) & 45.3 & 4.91 & 193.2 \\
Self-Compression ($K{=5}$) & 78.2 & 1.28 & 381.7 \\
\bottomrule
\end{tabular}
}
\caption{Ablation study on self-compression. \textit{SR}, \textit{Avg. $c_t$}, and \textit{Avg. Vis. Tok.} denote the success rate, average compression factor, and average vision tokens.
}
\label{tab:ablation}
\end{table}

At last, we conduct ablation studies to analyze the effectiveness of the agentic self-compression mechanism using the Qwen2.5-VL-3B-Instruct on the ALFWorld. The results are summarized in Tab.~\ref{tab:ablation}.

The results demonstrate that RL is essential for effectively leveraging the self-compression mechanism. In the absence of RL, \methodname{} lacks the prior knowledge required to modulate the compression factor, resulting in a negligible change in token usage and a slight performance decline compared to the fixed baseline. When RL is applied with a dense reward schedule ($K=1$), the agent prioritizes the immediate compression reward by aggressively increasing the compression factor to 4.91, which degrades visual fidelity and causes the success rate to plummet to 45.3\%. However, by adopting the intermittent reinforcement schedule ($K=5$), \methodname{} successfully balances the trade-off between information density and token cost. This configuration learns a favorable compression rate, reducing average visual token consumption from 458.1 to 381.7 while maintaining a success rate of 78.2\%, comparable to the 78.5\% achieved by the uncompressed visual baseline.



\section{Conclusions}

In this work, we present \methodname{} as an exploration into the potential of visual tokens as a compact history medium for multi-turn LLM agents. By integrating segment optical caching to mitigate rendering overhead and agentic self-compression to adaptively balance cost and fidelity, our method demonstrates that the visual modality can effectively complement textual history. Empirical results on ALFWorld and search-based QA suggest that this optical approach allows agents to retain the majority of their decision-making capabilities while significantly reducing token consumption, offering a resource-efficient alternative to text-only processing. We envision future research expanding on this foundation to explore hybrid storage architectures and unified multimodal interfaces, moving closer to the versatile and efficient information processing found in biological systems.

\section*{Limitations}
While \methodname{} demonstrates promising results in agentic tasks, several limitations warrant discussion and suggest directions for future work:

First, \methodname{} relies on off-the-shelf VLMs (Qwen2.5-VL series) that were not specifically designed for OCR intensive tasks. 
Although the proposed \methodname{} framework is model-agnostic in principle, we do not evaluate its behavior across a broader range of VLM architectures, like DeepSeek-OCR~\citep{deepseekocr}, with different visual tokenization strategies or patch resolutions. Performance and compression efficiency may vary depending on the inductive biases of the underlying vision encoder.

Second, \methodname{} relies on a deterministic text-to-image renderer with fixed hyperparameters such as font size, line spacing, color schemes, and image resolution. While we observe stable performance under our default configuration, we do not systematically explore the sensitivity of the agent to different rendering choices. Suboptimal rendering settings may reduce text legibility or distort layout cues, potentially affecting downstream reasoning.

Last, the current design assumes agent history consists primarily of text (observations, actions, tool outputs) that can be rendered as text-as-image. However, many realistic agent scenarios involve inherently visual elements: GUI screenshots with complex layouts, scientific plots, diagrams, and structured tables. Investigating compression strategies for such multimodal histories could significantly expand \methodname{}'s application scope beyond text-centric domains.

\section*{Ethical Considerations}
In this work, all experiments were conducted using established public benchmarks and synthesized environments, which do not involve any sensitive personal information or data privacy concerns. Since our work currently does not involve real-world deployment, its immediate social impact is limited. However, from a long-term perspective, the proposed self-compressing optical context significantly reduces the token overhead for agentic tasks, which may contribute to the development of energy-efficient "Green AI" and facilitate deployment on resource-constrained devices in the future.
We used LLMs as writing assistants to refine language. Specifically, their use is limited to grammar correction, style improvement, and wording adjustments for clarity and conciseness.

\section*{Acknowledgments}
This research is supported by the RIE2025 Industry Alignment Fund – Industry Collaboration Projects (IAF-ICP) (Award I2301E0026), administered by A*STAR, as well as supported by Alibaba Group and NTU Singapore through Alibaba-NTU Global e-Sustainability CorpLab (ANGEL).

\bibliography{custom}

\appendix

\newpage

\section{Pseudo Code}
\label{app:pseudo}
We provide the detailed pseudo code for \methodname{} to facilitate reproduction. 
Algorithm~\ref{alg:env_wrapper} handles the backend mechanics: it maintains the segment cache, executes the rendering tool with the requested compression $c_{t}$, and computes the efficiency-aware reward.
Algorithm~\ref{alg:agent_rollout} details the interaction loop, where the policy explicitly selects the tool parameter $c_{t}$ to control the resolution of the subsequent visual observation.

\begin{algorithm}[tb]
   \caption{Environment Wrapper of \methodname{}}
   \label{alg:env_wrapper}
   \small
\begin{algorithmic}[1]
   \STATE {\bfseries Input:} Reward param $\lambda$, Compression interval $K$.
   \STATE {\bfseries Internal State:} Memory $\mathcal{M}$, Cache $\mathcal{C}^{(e)}$, Renderer $\mathcal{R}$.
   
   \STATE \textbf{Function} $\text{Step}(\mathbf{a}_t, c_t, t)$
   \STATE \hspace{1em} \textcolor{gray}{// 1. Physical environment step}
   \STATE \hspace{1em} Execute $\mathbf{a}_t$, receive $\mathbf{o}_{t+1}$, $r_t$, and $\text{success\_flag}$
   \STATE \hspace{1em} $\mathcal{M}_{t+1} \leftarrow \mathcal{M}_t \cup \{(\mathbf{o}_{t+1}, \mathbf{a}_t)\}$
   
   \STATE \hspace{1em} \textcolor{gray}{// 2. Optical memory rendering (tool execution $\mathcal{R}$)}
   \STATE \hspace{1em} $\mathbf{h}_{t+1} \leftarrow \mathrm{Fetch}(\mathcal{M}_{t+1})$
   \STATE \hspace{1em} Segments $(\ell_{t+1,1}, \dots, \ell_{t+1,K}) \leftarrow \mathrm{Split}(\mathbf{h}_{t+1})$
   \STATE \hspace{1em} $\mathbf{L}_{\text{imgs}} \leftarrow []$
   \STATE \hspace{1em} \textbf{for} $i = 1$ \textbf{to} $K$ \textbf{do}
   \STATE \hspace{2em} $k_i \leftarrow \mathrm{Hash}(\ell_{t+1,i})$
   \STATE \hspace{2em} \textbf{if} $k_i \notin \mathcal{C}^{(e)}$ \textbf{then}
   \STATE \hspace{3em} $\mathcal{C}^{(e)}[k_i] \leftarrow \mathcal{R}(\ell_{t+1,i}; \psi)$ \COMMENT{Cache miss: render segment}
   \STATE \hspace{2em} \textbf{end if}
   \STATE \hspace{2em} Append $\mathcal{C}^{(e)}[k_i]$ to $\mathbf{L}_{\text{imgs}}$
   \STATE \hspace{1em} \textbf{end for}
   \STATE \hspace{1em} $\mathbf{I}_{\text{raw}} \leftarrow \mathrm{Stack}(\mathbf{L}_{\text{imgs}})$
   
   \STATE \hspace{1em} \textcolor{gray}{// Apply compression $c_t$ (Eq.~(\ref{eq:compress}))}
   \STATE \hspace{1em} $\mathbf{I}_{t+1} \leftarrow \mathrm{Resize}(\mathbf{I}_{\text{raw}}, \text{scale}=1/\sqrt{c_t})$
   
   \STATE \hspace{1em} \textcolor{gray}{// 3. Compression-aware reward (Eq.~(\ref{eq:compress_reward}))}
   \STATE \hspace{1em} \textbf{if} $\text{success\_flag}$ is True \textbf{then}
   \STATE \hspace{2em} $r^{\text{comp}}_t \leftarrow \ln(c_t)$ \, if success
   \STATE \hspace{1em} \textbf{else}
   \STATE \hspace{2em} $r^{\text{comp}}_t \leftarrow 0$
   \STATE \hspace{1em} \textbf{end if}
   
   \STATE \hspace{1em} \textcolor{gray}{// Apply sparse reward injection}
   \STATE \hspace{1em} $\widetilde{r}_t \leftarrow r_t + \lambda \cdot r^{\text{comp}}_t \cdot \mathbb{I}(t \mod K = 0)$
   
   \STATE \hspace{1em} {\bfseries return} $(\mathbf{o}_{t+1}, \mathbf{I}_{t+1})$, $\widetilde{r}_t$
   \STATE \textbf{End Function}
\end{algorithmic}
\end{algorithm}

\begin{algorithm}[tb]
   \caption{\methodname{} Policy Rollout}
   \label{alg:agent_rollout}
   \small
\begin{algorithmic}[1]
   \STATE {\bfseries Input:} Task Instruction $\mathcal{I}$, Total training steps $T_{\text{max}}$.
   \STATE {\bfseries Initialize:} Policy $\pi_\theta$, Env Wrapper $\mathcal{E}$.
   \STATE {\bfseries Initialize:} Global training step $t \leftarrow 0$.
   
   \WHILE{$t < T_{\text{max}}$}
       \STATE $\mathcal{E}.\text{Reset}()$
       \STATE Get initial observation $(\mathbf{o}_t, \mathbf{I}_t)$ from $\mathcal{E}$
       
       \WHILE{episode not done}
           \STATE \textcolor{gray}{// 1. Decision making (policy generates action + tool call)}
           \STATE Sample action $\mathbf{a}_t \sim \pi_\theta(\cdot \mid \mathcal{I}, \mathbf{I}_t)$
           
           \STATE \textcolor{gray}{// Parse specific tag as described in Sec.~\ref{sec:agentocr:self-compression}}
           \STATE Parse output $\rightarrow (\mathbf{a}_t, c_t)$ via <compression> tag
           
           \STATE \textcolor{gray}{// 2. Execute action and tool call}
           \STATE $(\mathbf{o}_{t+1}, \mathbf{I}_{t+1}), \widetilde{r}_t \leftarrow \mathcal{E}.\text{Step}(\mathbf{a}_t, c_t, t)$
           
           \STATE Store transition for optimization using $\widetilde{r}_t$
           \STATE $(\mathbf{o}_t, \mathbf{I}_t) \leftarrow (\mathbf{o}_{t+1}, \mathbf{I}_{t+1})$
           \STATE $t \leftarrow t + 1$
       \ENDWHILE
       
       \STATE Update $\pi_\theta$ via RL optimizer
   \ENDWHILE
\end{algorithmic}
\end{algorithm}

\section{Experiments}
\label{app:experiment}
\subsection{Details of Benchmarks}
\paragraph{ALFWorld.}
ALFWorld ~\citep{ALFWorld} is an embodied environment comprising 3,827 tasks, which is publicly available for non-commercial research purposes. The objective for LLM agents is to accomplish household tasks spanning six categories: Pick \& Place, Examine in Light (Look), Clean \& Place (Clean), Heat \& Place (Heat), Cool \& Place (Cool), and Pick Two \& Place. At each interaction step, the LLM agent selects an action based on the current observation and interaction history, then receives feedback from the environment to verify task completion.
\paragraph{Search-based QA.}
We utilize the QA dataset used in Search-R1~\citep{jin2025search}, which is publicly available for non-commercial research purposes. It contains two categories of benchmark datasets. The first category is single-hop question answering, which includes NQ ~\citep{kwiatkowski2019natural}, TriviaQA ~\citep{joshi2017triviaqa}, and PopQA ~\citep{mallen2022not}. The second category is multi-hop question answering, which includes HotpotQA ~\citep{yang2018hotpotqa}, 2WikiMultiHopQA (2Wiki) ~\citep{ho2020constructing}, MuSiQque ~\citep{trivedi2023interleaving}, and Bamboogle ~\citep{press2023measuring}. In this scenario, the agent autonomously generates search queries during step-by-step reasoning. It uses the E5 retriever~\citep{wang2022text} to retrieve relevant documents from a knowledge base, returning the top-3 most relevant passages for each query. The agent then analyzes and reasons over the retrieved information, iteratively refining its queries and incorporating new evidence until it arrives at the final answer.

\subsection{Details of Training}
\paragraph{Hyperparameters for ALFWorld.}
For ALFWorld, we follow the default settings used in GiGPO~\citep{feng2025group}. Text-only variants are assigned a maximum prompt length of 5120 tokens. Conversely, optical-history variants are constrained to a maximum prompt length of 2048 tokens. Across both modalities, the maximum response length is standardized to 512 tokens. The agent is permitted to interact with the environment for a maximum of 50 steps per episode, with full history. The learning rate is fixed at 1e-6. During the training phase, rollouts are executed on 16 samples per iteration, generating 8 trajectories per sample. We grant a reward of 10 for successful actions and 0 otherwise. The temperature parameter is set to 1.0 during rollouts and reduced to 0.4 for validation. The mini-batch size is maintained at 256. Consistent with standard practices, no KL-divergence loss is applied during training.

\paragraph{Hyperparameters for search-based QA.}
For the search-based QA tasks, text-only variants operate with an expanded maximum prompt length of 14000 tokens. For optical-history variants, the limit is set to 4096 tokens. Similar to ALFWorld, the maximum response length is capped at 512 tokens. The interaction horizon is shorter, allowing up to 4 environmental steps per episode and retaining a full history. The learning rate remains at 1e-6. Training rollouts involve 128 samples per iteration, with each sample producing 8 trajectories. The reward structure assigns 1 for correct answers and 0 for incorrect ones. Rollout temperature is set to 1.0, while validation uses a greedy decoding strategy (temperature 0.0). The mini-batch size is 256, and the KL-divergence loss is excluded.

\subsection{Computing Details}
All experiments utilizing the Qwen2.5-VL-3B-Instruct model for both ALFWorld and search-based QA were executed on 2×H100 GPUs. Conversely, the larger Qwen2.5-VL-7B-Instruct model required a configuration of 4×H100 GPUs. The agents were trained for a total duration of 150 iterations.

\subsection{Optical Rendering Details}
\label{app:rendering}

The optical memory is generated via a deterministic renderer $\mathbf{I}_t = \mathcal{R}(\mathbf{h}_t;\psi)$. We employ specific typographic settings and semantic color codes to facilitate efficient parsing by the VLM. The detailed rendering hyperparameters $\psi$ for both benchmarks are provided in Tab.~\ref{tab:rendering_params}.

\subsection{Prompts}
\begin{table}[h]
\centering
\resizebox{1.0\columnwidth}{!}{%
\begin{tabular}{lcc}
\toprule
\textbf{Parameter} & \textbf{ALFWorld} & \textbf{Search} \\
\midrule
\textit{Typography \& Layout} & & \\
\quad Font Family & Monospace & Monospace \\
\quad Font Size & 10pt & 12pt \\
\quad Line Spacing & 1.2 & 1.2 \\
\quad Max Width & 392px & 560px \\
\midrule
\textit{Semantic Color Mapping} & & \\
\quad Task \& Context & Black & Black \\
\quad [Observation] & Blue (0,0,255) & -- \\
\quad [Action] & Red (255,0,0) & -- \\
\quad <search> & -- & Blue (0,0,255) \\
\quad <information> & -- & Red (255,0,0) \\
\bottomrule
\end{tabular}%
}
\caption{Rendering hyperparameters $\psi$ for optical memory generation.}
\label{tab:rendering_params}
\end{table}
The specific prompts employed for agents on the ALFWorld and search-based QA tasks are illustrated in Fig.~\ref{prompt:alfworld_prompt_temp_text}, \ref{prompt:alfworld_prompt_temp_image}, \ref{prompt:serach_prompt_text}, and \ref{prompt:serach_prompt_image}. Specifically, Fig.~\ref{prompt:alfworld_prompt_temp_text} and \ref{prompt:serach_prompt_text} detail the templates for text-only baselines, whereas Fig.~\ref{prompt:alfworld_prompt_temp_image} and \ref{prompt:serach_prompt_image} present the templates for optical-history variants.

These templates are constructed using Python-style string formatting, where placeholders in curly braces (\{\}) mark semantic slots. For instance, \textcolor{brown}{\{task\_description\}} denotes the task definition, and \textcolor{brown}{\{current\_observation\}} indicates the immediate environmental feedback. These slots are populated with dynamic content during training.

To structure the model's reasoning and outputs, we utilize specific control tags. The \textcolor{DeepGreen}{<think>} \textcolor{DeepGreen}{</think>} tags enclose the mandatory step-by-step reasoning chain. Final decisions are wrapped within \textcolor{DeepPurple}{<action>} \textcolor{DeepPurple}{</action>} tags. In the context of search agents, queries are issued between \textcolor{DeepPurple}{<search>} \textcolor{DeepPurple}{</search>} tags, with retrieved evidence presented inside \textcolor{DeepBlue}{<information>} \textcolor{DeepBlue}{</information>} tags; final answers are enclosed in \textcolor{DeepPurple}{<answer>} \textcolor{DeepPurple}{</answer>}. Uniquely for our \methodname{} method, the model is instructed to specify the compression ratio using \textcolor{DeepOrange}{<compression>} \textcolor{DeepOrange}{</compression>} tags.

\begin{figure*}[t]
\centering
\resizebox{0.85\textwidth}{!}{
\begin{tcolorbox}[colback=gray!5!white, colframe=black!75!black, 
title=Prompt Template of Text Agent on ALFWorld, boxrule=0.3mm, width=\textwidth, arc=3mm, auto outer arc=true]
You are an expert agent operating in the ALFRED embodied Environment. Your task is to: \textcolor{brown}{\{task\_description\}}. Prior to this step, you have already taken \textcolor{brown}{\{step\_count\}} step(s). Below are the most recent \textcolor{brown}{\{history\_length\}} observations and the corresponding actions you took: \textcolor{brown}{\{action\_history\}}. You are now at step \textcolor{brown}{\{current\_step\}} and your current observation is: \textcolor{brown}{\{current\_observation\}}. Your admissible actions of the current situation are: [\textcolor{brown}{\{admissible\_actions\}}].

Now it's your turn to take an action.
You should first reason step-by-step about the current situation. This reasoning process MUST be enclosed within \textcolor{DeepGreen}{<think>} \textcolor{DeepGreen}{</think>} tags.
Once you've finished your reasoning, you should choose an admissible action for current step and present it within \textcolor{DeepPurple}{<action>} \textcolor{DeepPurple}{</action>} tags.
\end{tcolorbox}
}
\caption{The prompt template of text agent on ALFWorld.}
\label{prompt:alfworld_prompt_temp_text}
\end{figure*}

\begin{figure*}[t]
\centering
\resizebox{0.85\textwidth}{!}{
\begin{tcolorbox}[colback=gray!5!white, colframe=black!75!black, 
title=Prompt Template of \methodname{} on ALFWorld, boxrule=0.3mm, width=\textwidth, arc=3mm, auto outer arc=true]
<image>

You are an expert agent operating in the ALFRED embodied Environment. Your task is to: \textcolor{brown}{\{task\_description\}}. Prior to this step, you have already taken \textcolor{brown}{\{step\_count\}} step(s). The provided image shows the most recent \textcolor{brown}{\{history\_length\}} observations and the corresponding actions you took. You are now at step \textcolor{brown}{\{current\_step\}} and your current observation is: \textcolor{brown}{\{current\_observation\}}. Your admissible actions of the current situation are: [\textcolor{brown}{\{admissible\_actions\}}].

Now it's your turn to take an action.
You should first reason step-by-step about the current situation. This reasoning process MUST be enclosed within \textcolor{DeepGreen}{<think>} \textcolor{DeepGreen}{</think>} tags.
Once you've finished your reasoning, you should choose an admissible action for current step and present it within \textcolor{DeepPurple}{<action>} \textcolor{DeepPurple}{</action>} tags.

Additionally, select an image compression factor larger than 1.0 for the next image. Higher compression lowers cost, but too much compression harms image quality. You must provide the next compression factor within \textcolor{DeepOrange}{<compression>} \textcolor{DeepOrange}{</compression>} tags (e.g., \textcolor{DeepOrange}{<compression>}1.1\textcolor{DeepOrange}{</compression>}).
\end{tcolorbox}
}
\caption{The prompt template of \methodname{} on ALFWorld.}
\label{prompt:alfworld_prompt_temp_image}
\end{figure*}

\begin{figure*}[t]
\centering
\resizebox{0.85\textwidth}{!}{
\begin{tcolorbox}[colback=gray!5!white, colframe=black!75!black, 
title=Prompt Template of Text Agent on Search-based QA, boxrule=0.3mm, width=\textwidth, arc=3mm, auto outer arc=true]
You are an expert agent tasked with answering the given question step-by-step. 

Your question: \textcolor{brown}{\{task\_description\}}. 

Prior to this step, you have already taken \textcolor{brown}{\{step\_count\}} step(s). Below is the interaction history, where \textcolor{DeepPurple}{<search>}...\textcolor{DeepPurple}{</search>} wrapped your past search queries and \textcolor{DeepBlue}{<information>}...\textcolor{DeepBlue}{</information>} wrapped the corresponding search results. History: \textcolor{brown}{\{memory\_context\}}

Now it's your turn to respond for the current step.
You should first conduct a reasoning process. After completing your reasoning, choose only one of the following actions (do not perform both):

(1) If any required knowledge is missing or uncertain, you MUST call a search engine to get more external information using format: \textcolor{DeepPurple}{<search>} your query \textcolor{DeepPurple}{</search>}.

(2) Only if you have sufficient information to answer the question with high confidence, provide your final answer within \textcolor{DeepPurple}{<answer>} \textcolor{DeepPurple}{</answer>} tags.
\end{tcolorbox}
}
\caption{The prompt template of text agent on search-based QA.}
\label{prompt:serach_prompt_text}
\end{figure*}

\begin{figure*}[t]
\centering
\resizebox{0.85\textwidth}{!}{
\begin{tcolorbox}[colback=gray!5!white, colframe=black!75!black, 
title=Prompt Template of \methodname{} on Search-based QA, boxrule=0.3mm, width=\textwidth, arc=3mm, auto outer arc=true]
<image>

You are an expert agent tasked with answering the given question step-by-step. 

Your question: \textcolor{brown}{\{task\_description\}}. 

Prior to this step, you have already taken \textcolor{brown}{\{step\_count\}} step(s).
The image contains the full history:

- Past queries are inside \textcolor{DeepPurple}{<search>}...\textcolor{DeepPurple}{</search>}

- Past results are inside \textcolor{DeepBlue}{<information>}...\textcolor{DeepBlue}{</information>}

Now it's your turn to respond for the current step.
You should first conduct a reasoning process. After completing your reasoning, choose only one of the following actions (do not perform both):

(1) If any required knowledge is missing or uncertain, you MUST call a search engine to get more external information using format: \textcolor{DeepPurple}{<search>} your query \textcolor{DeepPurple}{</search>}.

(2) Only if you have sufficient information to answer the question with high confidence, provide your final answer within \textcolor{DeepPurple}{<answer>} \textcolor{DeepPurple}{</answer>} tags.

Additionally, select an image compression factor larger than 1.0 for the next image. Higher compression lowers cost, but too much compression harms image quality. You must provide the next compression factor within \textcolor{DeepOrange}{<compression>} \textcolor{DeepOrange}{</compression>} tags (e.g., \textcolor{DeepOrange}{<compression>}1.1\textcolor{DeepOrange}{</compression>}).

Output format:

1. Reasoning: state what you found in the image.

2. \textcolor{DeepPurple}{<search>}...\textcolor{DeepPurple}{</search>} or \textcolor{DeepPurple}{<answer>}...\textcolor{DeepPurple}{</answer>}

3. \textcolor{DeepOrange}{<compression>}...\textcolor{DeepOrange}{</compression>}
\end{tcolorbox}
}
\caption{The prompt template of \methodname{} on search-based QA.}
\label{prompt:serach_prompt_image}
\end{figure*}

\section{Case Study}
\begin{figure*}
    \centering
    \includegraphics[width=0.6\linewidth]{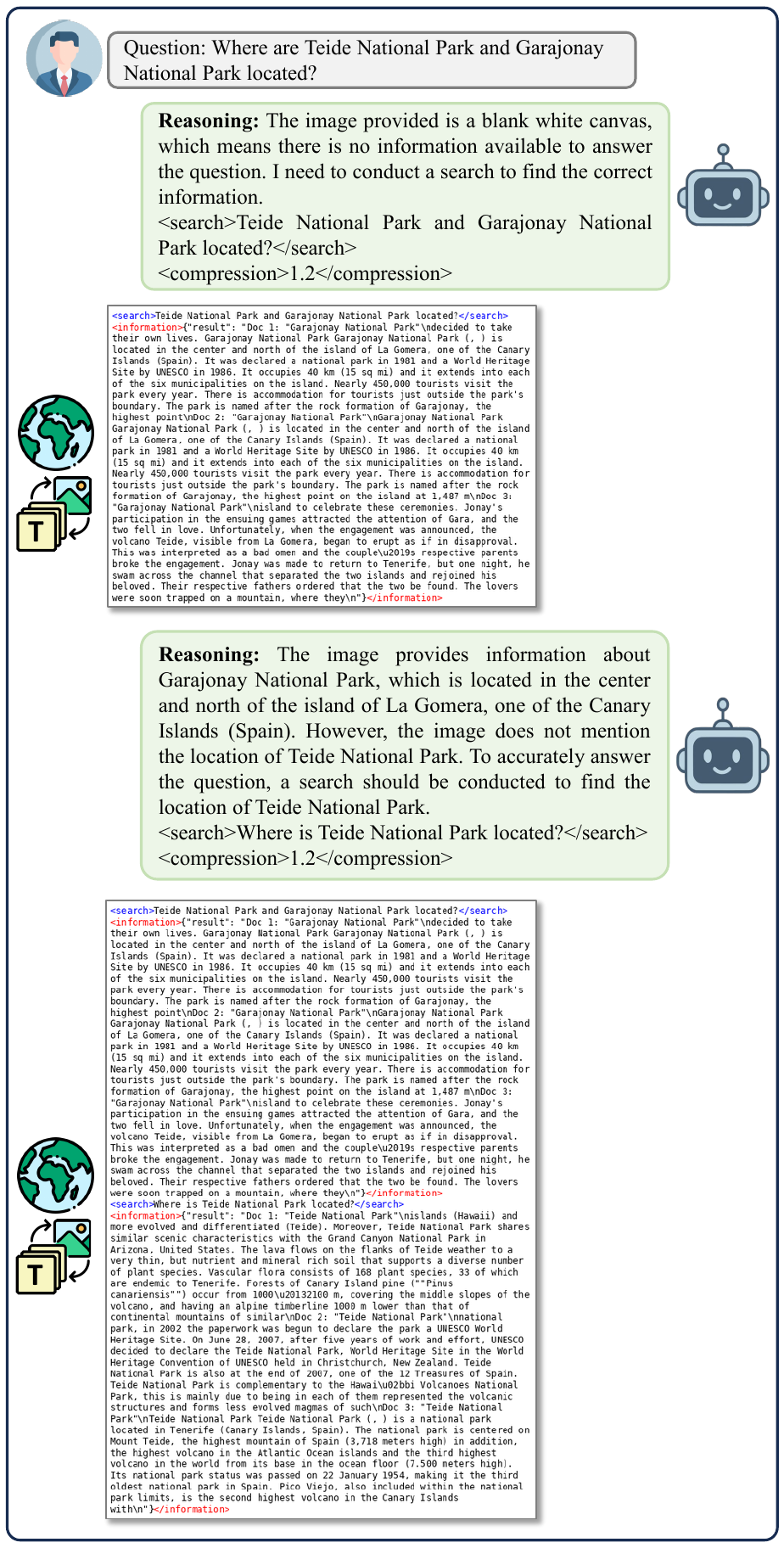}
    \caption{Case study of \methodname{} on HotpotQA (part I)}
    \label{fig:case_study_part1}
\end{figure*}

\begin{figure*}
    \centering
    \includegraphics[width=0.6\linewidth]{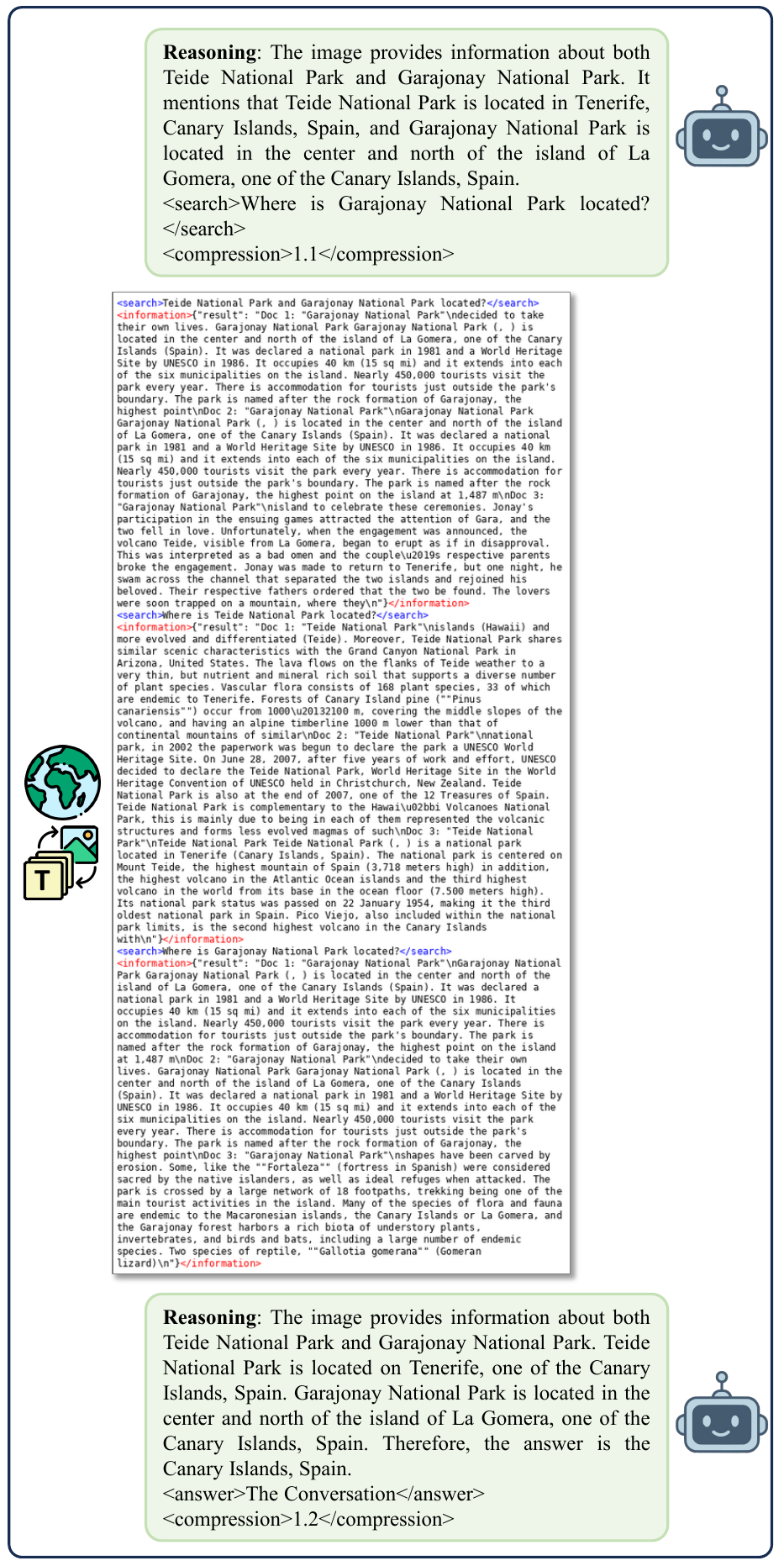}
    \caption{Case study of \methodname{} on HotpotQA (part II)}
    \label{fig:case_study_part2}
\end{figure*}

We present the complete trajectory of multi-turn interaction between the agent and the search tool, and demonstrate the reasoning behavior of \methodname{} trained by GRPO. As shown in Fig.~\ref{fig:case_study_part1} and Fig.~\ref{fig:case_study_part2}, the agent progressively accumulates search results in its optical memory and adaptively adjusts compression factors at each step. The agent successfully retrieves relevant information across multiple search iterations and arrives at the correct answer while maintaining efficient token usage through visual history compression.

\end{document}